\documentclass[conference]{IEEEtran}
\usepackage{times}
\usepackage{caption}
\usepackage{subcaption}

\usepackage[numbers]{natbib}
\usepackage{multicol}
\usepackage[bookmarks=true]{hyperref}
\usepackage{amsfonts}
\usepackage{amsmath}
\usepackage{comment}
\usepackage{graphicx}
\usepackage{algorithm}
\usepackage{algorithmic}
\usepackage{xcolor}
\usepackage{booktabs} 
\usepackage{threeparttable}
\usepackage{xspace}
\title{A low-cost and lightweight 6 DoF bimanual arm \\ for dynamic and contact-rich manipulation}

\let\oldtwocolumn\twocolumn
\renewcommand\twocolumn[1][]{%
    \oldtwocolumn[{#1}{
    \begin{center}
           \includegraphics[width=0.9\textwidth]{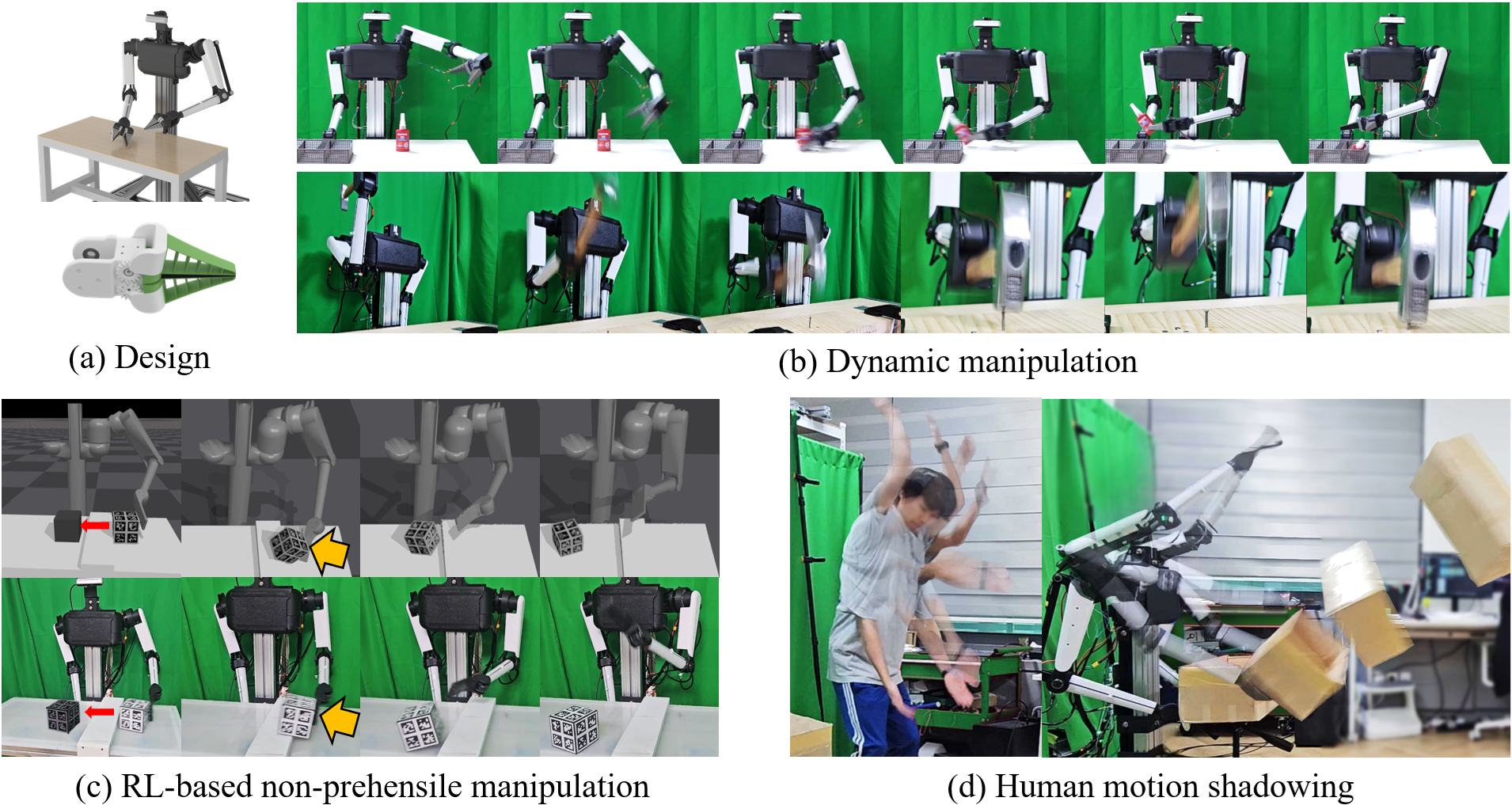}
           \captionof{figure}{(a) Illustration of the ARMADA design. (b) ARMADA's lightweight 6-DoF bimanual arms enable dynamic manipulations such as object snatching and hammering. (c) The platform supports sim-to-real transfer of dynamic motions using reinforcement learning policies. (d) ARMADA also demonstrates the ability to shadow dynamic human movements.}
           \label{fig:overall_illustration}
        \end{center}
    }]
}

\newcommand{\robot}{{ARMADA}\xspace} 

\begin{document}





%
\author{
\authorblockN{Jaehyung Kim\authorrefmark{1},
Jiho Kim\authorrefmark{2},
Yujin Jang\authorrefmark{2}, 
Dongryung Lee\authorrefmark{1} and
Beomjoon Kim\authorrefmark{1}}
\authorblockA{\authorrefmark{1}Kim Jaechul Graduate School of AI, KAIST}
\authorblockA{\authorrefmark{2}Department of Mechanical System Design Engineering, SEOULTECH}
}

\maketitle
\thispagestyle{empty}
\pagestyle{empty}

\begin{abstract}
Dynamic and contact-rich object manipulation, such as striking, snatching, or hammering, remains challenging for robotic systems due to hardware limitations. Most existing robots are constrained by high-inertia design, limited compliance, and reliance on expensive torque sensors. To address this, we introduce ARMADA (Affordable Robot for Manipulation and Dynamic Actions), a 6 degrees-of-freedom bimanual robot designed for dynamic manipulation research. ARMADA combines low-inertia, back-drivable actuators with a lightweight design, using readily available components and 3D-printed links for ease of assembly in research labs.  The entire system, including both arms, is built for just \$6,100. Each arm achieves speeds up to 6.16m/s, almost twice that of most collaborative robots, with a comparable payload of 2.5kg. We demonstrate ARMADA can perform dynamic manipulation like snatching, hammering, and bimanual throwing in real-world environments. We also showcase its effectiveness in reinforcement learning (RL) by training a non-prehensile manipulation policy in simulation and transferring it zero-shot to the real world, as well as human motion shadowing for dynamic bimanual object throwing. ARMADA is fully open-sourced with detailed assembly instructions, CAD models, URDFs, simulation, and learning codes at \it{\url{https://sites.google.com/view/im2-humanoid-arm}}.

\end{abstract}

\IEEEpeerreviewmaketitle

\section{Introduction}

Humans use a rich set of action repertoire to manipulate objects: we not only pick-and-place objects but also toss laundry, slide a box, snatch a pen, hammer a nail, otherwise leverage momentum and forces to manipulate diverse objects efficiently and effectively. In contrast, most manipulators today are limited to picking, where a robot simply grasps an object to resist the frictional force. While kinematic pick-and-place is sufficient for static, controlled tasks, dynamic manipulation is necessary for building an effective general-purpose robot.

One critical reason for the lack of dynamic manipulation capability is hardware. Traditional industrial robots are strong, precise, and consistent, but their heavy and high inertia structures make them unsuitable for dynamic manipulation in human environments. Modern collaborative robots, such as the Franka Panda~\cite{haddadin2022franka}, are designed to work alongside humans by using a lighter and smaller design. However, they still have high inertia and are constrained by velocity and torque limits to ensure safety, rendering them inadequate for dynamic manipulation that requires high speed and acceleration. Furthermore, they use high gear ratio actuators which effectively ``lock'' joints, making them difficult to absorb high impact forces at contacts. Such actuators also require expensive torque sensors, as current-to-torque relationship is difficult to model due to the backlash and friction caused by multi-stage gears.

Our objective is to build and open-source a 6 degrees-of-freedom (DoF) bimanual robot that can be easily and cheaply assembled at a lab to democratize dynamic manipulation research, such as examples shown in Figure~\ref{fig:overall_illustration}. To achieve this, our design must meet the following criteria:

\begin{itemize}
    \item \textbf{Dynamic}: the robot should be able to move at high speed with acceleration to perform dynamic manipulation.
    \item \textbf{Low-cost}: the design should use inexpensive materials, sensors, and actuators while having high payload, precision, and torque necessary for versatile manipulation.
    \item \textbf{Safe}: the robot should have a low inertia to ensure safety even at the maximum speed.
    \item \textbf{Ease of assembly}: the robot should be built with off-the-shelf materials.
\end{itemize}


One potential way to achieve these is through tendon drive systems~\cite{song2018development, quigley2011low, guist2024safe}. These systems transmit torques to joints located far from motors, allowing actuators to be placed at the arm's base. This design significantly reduces inertia and enhances the robot's agility. Furthermore, tendons are highly back-drivable, making them well-suited for rapid and dynamic movements.

However, tendon-driven robots have several problems. First, they are difficult to assemble and maintain at a research lab. Assembling a tendon-wiring mechanism involves integrating multiple pulleys connecting tendons to rotors while avoiding interference with electronic connections, an intricate and complex task. Achieving the desired levels of stiffness and friction adds another layer of complexity, as it requires precise adjustment of tendon tension through a tensioner. Moreover, tendon-driven systems are susceptible to wear and tear, particularly during high-impact tasks such as hammering. In such scenarios, tendons may loosen upon impact, necessitating frequent adjustments and repairs to maintain performance.

We instead take inspiration from the recent success of quadrupeds in designing our bimanual robot. Recent quadrupeds demonstrate dynamic and explosive movements, such as jumping and parkour~\cite{hoeller2024anymal, cheng2024extreme}, while reliably withstanding high-impact, high-frequency contact forces. What enables this is the use of quasi-direct drive (QDD) actuators, which have significantly lower gear ratios compared to those typically used in collaborative robots (eg., 1:10 vs. 1:100). These low-gear ratio actuators make the system highly back-drivable, allowing joints to absorb impact forces through natural compliance. Additionally, the reduced backlash and friction in these actuators simplify modeling the current-to-torque relationship, eliminating the need for expensive torque sensors.

To minimize inertia, we would ideally move all the actuators to the body, as done in quadrupeds. However, the key difference between a quadruped leg and a bimanual arm is the number of joints. While a quadruped leg typically has 3 DoF, with only the knee joint located away from the body, a robot arm requires at least 6 DoF to achieve full spatial manipulation. To address this, we mount heavier and stronger actuators on the main body to control the shoulder joints, and use smaller, lighter actuators for the elbow and wrist joints. This design results in a lighter moving mass, albeit at the cost of reduced strength in the elbow and wrist.

Figure~\ref{fig:overall_illustration}a top showcases our robot, \robot (Affordable Robot for Manipulation and Dynamic Actions), built with these design principles. Each arm weighs 1.09 kg (excluding body-mounted parts and gripper) and is constructed using off-the-shelf motors and 3D-printed links. The entire system costs \$6,100 to build. To eliminate the need for a torque sensor at each joint, we manually measured and calibrated the current-to-torque relationship\footnote{vendor's data was inaccurate}. To improve impact resistance and ease of assembly, we use a linkage-based transmission mechanism that does not require tensioners. The robot’s links are 3D-printed using polylactic acid (PLA), except for the elbow joint, which is made from aluminum to reduce deformation under high loads. 

We also develop a simple, compact jaw gripper compatible with the arm shown in Figure~\ref{fig:overall_illustration}a bottom.  By using the thermoplastic polyurethane (TPU) based finger without linkage structure, this design simplifies construction and maintenance while providing sufficient robustness for dynamic tasks. The gripper design, like the rest of \robot, is fully open-sourced.

In our experiments, we show \robot could perform several dynamic motions, such as object snatching and hammering. We also demonstrate we can train a contact-rich non-prehensile manipulation policy entirely in simulation using reinforcement learning (RL), and zero-shot transfer to the real world. Lastly, we show \robot can be used for human motion retargeting on the dynamic bimanual object throwing task. Examples from these tasks are highlighted in Figure~\ref{fig:overall_illustration}. We completely open-source our code and design.

\section{Related Work}

\subsection{Existing manipulators in the context of dynamic manipulation}
\label{sec:existing_manipulators}

\begin{table*}[ht]
\resizebox{\textwidth}{!}{%
\centering
\begin{threeparttable}
\caption{Comparison of manipulators}
\label{tab:comparison}
\begin{tabular}{l||c|c|c|c|c|c|c|c}

                                  & \robot (Ours)    & Franka Panda~\cite{haddadin2022franka} & KUKA iiwa 7 R800 & Quigley et al.~\cite{quigley2011low} & LIMS~\cite{kim2017anthropomorphic}    & Nishii et al.~\cite{nishii2023ultra} & PAMY2~\cite{guist2024safe}    & BLUE~\cite{gealy2019quasi}              \\ \hline
DoF (one arm)                 & 6       & 7            & 7                & 7              & 7       & 6                      &  {4}        & 7\\ 
Inertia (kg\ensuremath{\cdot}m\textsuperscript{2})            &  {0.234}   &  {large}           &  {large}               &  {0.083}        &  {0.599}   & ?       & ?        &  {0.75}\\ 
Moving mass\tnote{1} (kg)   &  {1.09}    &  {18}           &  {22.3}             &  {2}              &  {2.24}    & {0.176}   &  {1.3}      &  {8.7}\\ 
End-effector speed (m/s)    &  {6.16}    &  {1.7}          & 3.2              &  {1.5}            &  {5.35}    & ?       &  {12}       & 2.1\\ 
Total cost (\$, one arm)      &  {3,040} &  {expensive}           &  {expensive}               &  {4,135}          & ?       & ?       &  {14,540}     &  {\textless{}5,000} \\ 
Open-source                   &  {O}       &  {X}            &  {X}                &  {O}              &  {X}       &  {X}        &  {O}        &  {O}                 \\ 
Payload (kg)                & 2.5     & 3            & 7                & 2              & 3       & 3                      & ?        & 2\\ 
\end{tabular}
\begin{tablenotes}
\item ``?" denotes information not provided in the paper.
\item[1] moving mass is defined as the arm’s mass, excluding body-mounted components and the gripper.
\end{tablenotes}
\end{threeparttable}
}
\end{table*}

Collaborative robots (cobots), such as the Franka Emika~\cite{haddadin2022franka} and KUKA LBR series~\cite{hirzinger2001new, hirzinger2002dlr, albu2007dlr}, are designed to operate in humans environments. One significant limitation of cobots is their reliance on high-gear-ratio actuators, which introduce uncertainties in joint dynamics due to gearbox backlash and friction. Because of this, to achieve precise torque control, cobots often incorporate torque sensors; however, these sensors are susceptible to damage from impacts, and are expensive~\cite{zhu2023design}. Furthermore, while cobots are smaller in size, they still have the same design structure as industrial robots, with heavy actuators (around 2 kg) placed at each joint, which results in a high-inertia arm.

The dominant approach for designing robots for dynamic manipulation is to use tendon transmissions to achieve low inertia and absorb impact forces. Barrett WAM~\cite{rooks2006harmonious} uses a cable-and-cylinder drive system to mount four motors for the shoulder and elbow joints on the robot’s body, resulting in a low-inertia 4 DoF arm capable of tasks such as batting and throwing. Similarly, Quigley et al.~\cite{quigley2011low} incorporate series-elastic actuators at the shoulder and elbow joints with tendon-based transmission to achieve low inertia. LIMS~\cite{kim2015design, song2018development} attach all seven actuators to the main body and control joints via tendons. Coupled with lightweight link designs, it can perform dynamic tasks such as swinging folding fans. PAMY2~\cite{guist2024safe} combines tendon-driven transmissions with pneumatic artificial muscles for passive compliance in a 4 DoF arm, demonstrating dynamic capabilities such as table tennis smashing. Nishii et al.~\cite{nishii2023ultra} uses a combination of timing belts and tendons. It uses a timing belt to attach four motors near the body, and a tendon-based quaternion joint mechanism for the 2 DoF wrist, achieving 6 DoF maneuverability with reduced moving mass. 

Tendon-based systems generally achieve low inertia and gear ratio, and enable quick, explosive, and dynamic movements. However, maintaining tendons requires precise tensioning to achieve proper stiffness and friction, and it is very difficult to assemble them. These systems also face wear and tear during high-impact tasks, requiring frequent repairs. We summarize the comparison among \robot and existing manipulators in Table~\ref{tab:comparison}.

\subsection{Dynamic movements in quadrupeds}
\begin{figure}[ht]
\centering
\includegraphics[width=0.8\linewidth]{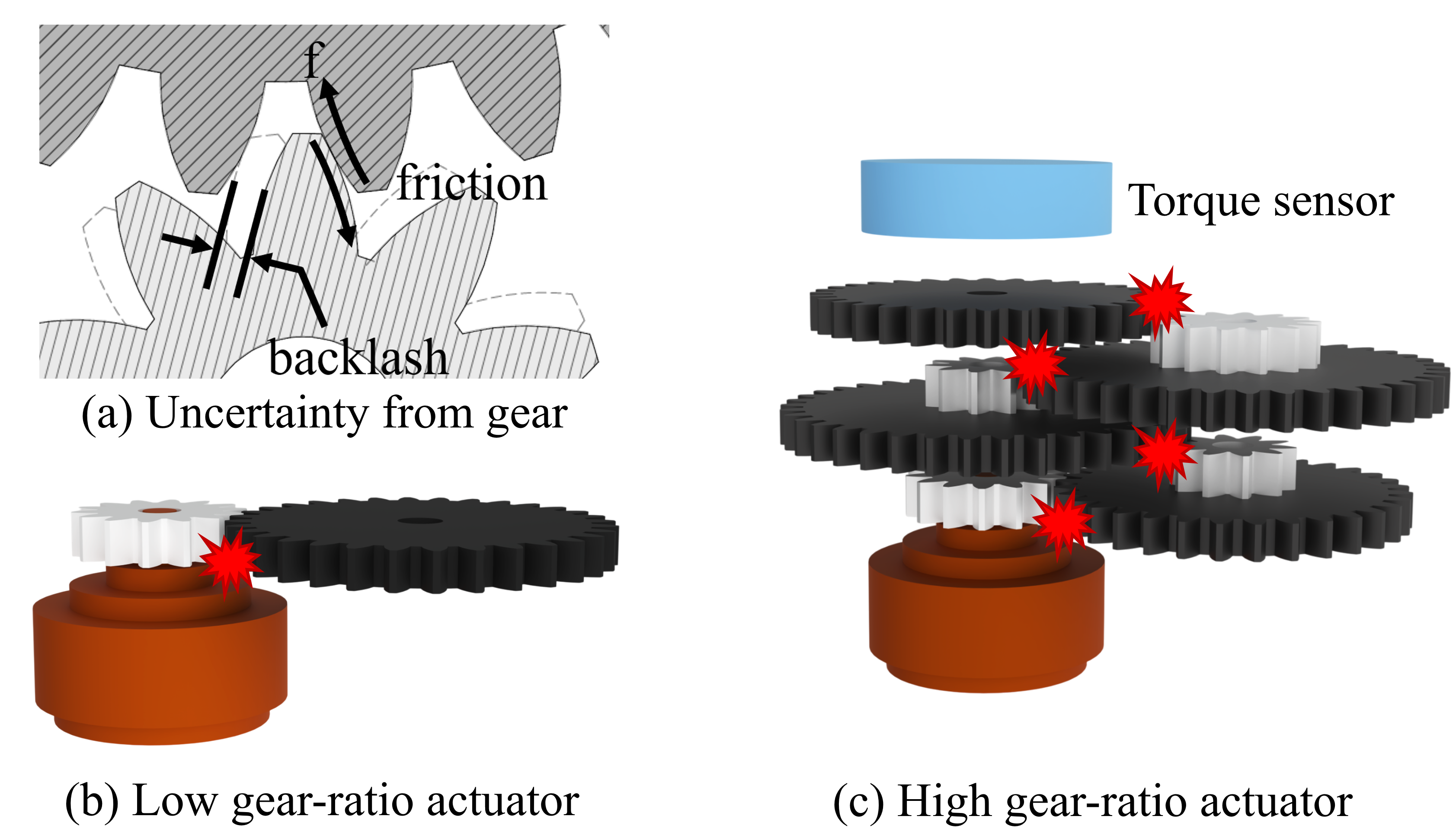}
\caption{Comparison of gearboxes in low-gear ratio actuators (b) vs. high-gear ratio actuators (c).  All gear mechanisms introduce backlash and friction (a) which are difficult to model. Because low gear ratios use fewer gears, it is easier to model.}
\label{fig:gear_uncertainty}
\end{figure}

The state-of-the-art quadruped hardware, combined reinforcement learning, has shown impressive dynamic motions such as jumping and parkour recently~\cite{zhuang2023robot, cheng2024extreme, choi2023learning}. Representative quadruped hardware includes, but is not limited to, Unitree Go1, MIT Cheetah~\cite{bledt2018cheetah}, KAIST Hound~\cite{shin2022design}, and Raibo~\cite{choi2023learning}.
One common factor in all these hardware platforms is the use of quasi-direct-drive (QDD) actuators~\cite{wensing2017proprioceptive}. Since QDD actuators 
have fewer gear stages, they have much less backlash and friction compared to high-gear ratio actuators (See Figure~\ref{fig:gear_uncertainty}). Further, their back-drivability makes them naturally compliant, enabling them to absorb high-impact forces, and allows us to model the current-to-torque relationship much easier. We take inspiration from these successes, and adopt QDD in our arms.

Another essential design principle in recent quadrupeds is the use of low-inertia legs, typically achieved by mounting actuators on the body and connecting them to the legs via transmission systems such as timing belts or linkages. This approach minimizes the moving mass, enabling the legs to move rapidly while using less torque. In our arm, we also employ linkages as our transmission system to control the elbow joint since it is easier to assemble, and locate as many actuators as possible to the robot's body. However, unlike quadrupeds, our arm features six degrees of freedom (6 DoF) and requires additional degrees of freedom at the wrist. To address this and maintain low inertia, we use lighter, weaker actuators at the wrist since high torque is generally not required in the regions of these parts.

\subsection{Hardware solutions for high-impact manipulation}

For dynamic and contact-rich tasks, robustness against impact forces is essential. Several works have proposed hardware solutions to address this in high-impact scenarios such as hammering. Izumi et al.~\cite{izumi1993control} utilize a flexible link to minimize impulsive forces, and Garabini et al.~\cite{garabini2011optimality} use actuators with torsional elastic springs to absorb impact while achieving high speeds. 
Both approaches model the system to control and demonstrate the hammering task but require accounting for damping forces and vibrations from elastic components, making them difficult to simulate.
Instead of mitigating impact forces with elastic materials, \robot adopts QDD actuators to make the system inherently compliant which simplifies modeling and simulation.

Other designs~\cite{imran2016impulse, romanyuk2019multiple} focus on adding links or control modes to manage impact forces for hammering. Imran et al.~\cite{imran2016impulse} introduce serial linkage chains around the robot links to distribute impact forces, while Romanyuk et al.~\cite{romanyuk2019multiple} employ MRR actuators~\cite{liu2008development} with multiple working modes. By switching to a passive mode during impacts, these actuators allow free rotation to absorb impulsive forces. Although these design choices effectively absorb impacts, additional links or control modes complicate the robot’s design, making them harder to assemble and control. Instead of increasing complexity, \robot uses a simpler design with just a four-bar linkage and QDD actuators, ensuring ease of assembly and control while maintaining robustness to impacts.

\subsection{Dynamic manipulation with existing manipulators}
There have been several attempts to generate dynamic manipulation motions using existing arms, although they have been limited to one specific category of motion such as throwing, catching, or batting. For throwing, Senoo et al.~\cite{high_speed_throwing} models contact at the robot fingertip to throw with the WAM robot (4 DoF). Bombile and Billard~\cite{kuka_grabbing_tossing} formulate modulated dynamical systems of a pair of KUKA LBR to toss a box with dual manipulators. Recently, some works using learning to account for the physical properties not explicitly considered in the projectile mechanics from perceptual input, using a Tx60 (7 DoF)~\cite{throwing_visual_feedback}, a single PR2 arm (7 DoF)~\cite{dppt}, and UR5 (6 DoF)~\cite{zeng2020tossingbot}. 

For catching, Lampariello et al.~\cite{planning_for_catching}  model the motion of 7 DoF manipulator (KUKA LBR) as a dynamical system and propose real-time online optimization algorithms to catch an object tracked with a vision-based motion capture system. Kim et al.~\cite{kim2014catching} leverage expert demonstration to guide the same 7 DoF manipulator to catch more diverse objects.  B\"{a}tz et al.~\cite{batz2010dynamic} predicts the trajectory of a flying ball and plausible interception points computed offline to catch the ball with Staubli RX90B (6 DoF). 
These works focus on the light object which induces negligible impulse since the aforementioned manipulators lack compliance to cope with the large impact generated during the catching of heavier objects. Salehian et al.~\cite{salehian2016dynamical} and Yan et al.~\cite{yan2024impact} attempt to mitigate the impact in the non-compliant manipulator (KUKA LBR) by separately planning a decelerating motion after the intercept of the object.  However, additional computation is required for such motion, which often results in failing to plan the motion before the object falls. 
Unlike these robots, \robot~is naturally compliant and can avoid such additional computation.  Similarly, Kim et al.~\cite{free_flying} enables fast catching with a compliant Barret WAM (7 DoF), yet maintenance of the robot is difficult due to its tendon-based mechanism.

For batting, Senoo et al.~\cite{high_speed_batting} develop a high-frequency vision system to track a ball and bat it with a Barrett WAM (4 DoF) using handcrafted batting dynamics. Similarly, Jia et al.~\cite{jia2019batting} model 2D impact dynamics with Coloumb friction and energy-based restitution to bat an object to a desired position with the same 4 DoF manipulator. Although these methods enable dynamic batting with light objects like styrofoam or ping pong balls, it is difficult to use them with heavier objects with existing arms, as hitting heavy objects might damage the manipulator. In contrast, \robot~is naturally compliant, and is easy to maintain.

\subsection{RL-based Manipulation}


Recently, reinforcement learning (RL) has been successfully applied to contact-rich manipulation tasks, including in-hand manipulation~\cite{chen2022system, andrychowicz2020learning, allshire2022transferring, handa2023dextreme} and non-prehensile manipulation~\cite{zhou2023hacman, zhou2023learning, cho2024corn, kim2023pre}, addressing challenges that traditional planning techniques have struggled to solve. These advancements have been made possible through the combination of large-scale simulation and simulation-to-reality transfer techniques, such as domain randomization. However, most of these works rely on cobots, which are less suitable for quick and dynamic motions involving frequent impacts limiting them to slow movements. In contrast, we demonstrate that our robot can achieve similar non-prehensile manipulation tasks using RL while exhibiting much more agile and dynamic motions.

\subsection{Human motion shadowing}

Thanks to the recent development of computer vision algorithms, several works show that we can collect human demonstrations simply by watching them. H2O~\cite{h2o} and OmniH20~\cite{omnih2o} adopts HybrIK~\cite{hybrik} as body pose predictors to achieve real-time humanoid teleoperation using a RGB camera. HumanPlus~\cite{humanplus} leverages human body~\cite{wham} and hand~\cite{hamer} pose predictors to extract human joint angles in real time and project them on the humanoid actuators. Similarly, OKAMI~\cite{okami} adopts body~\cite{slahmr} and hand~\cite{hamer} pose predictors to extract human body keypoints and solve differential inverse kinematics~\cite{pink} to imitate human motion that keeps the relative positions between the hands similar to that of human. Leveraging a single RGB camera with body pose predictor~\cite{wham} we showcase that our hardware, \robot, also is capable of shadowing human demonstrations in dynamic bimanual throwing.

\section{Design of \robot}

This section describes our hardware design and implementation details. Our primary goal is to achieve high-speed manipulation with low gear-ratio proprioceptive actuators while maintaining sufficient durability and ease of maintenance.

\subsection{Proprioceptive actuators}

\subsubsection{Choice of actuators}
We choose two types of off-the-shelf proprioceptive actuators: T-motor AK70-10 for shoulder and elbow and Cubemars RMD X4 V2 for wrist and forearm. Both types have 1:10 gear ratio with integrated drivers. Each arm is powered by two 24V Mean Well LRS-600-24 switch mode power supplies (SMPS). To protect the power supply from the back electromotive force generated by the actuators, SEMI-REX MD110-16 diodes are installed between the power supply and the actuators. 

\subsubsection{Sensor-free torque estimation for control}
Instead of using expensive torque sensors, we estimate the torque output from the actuators by building a custom current-to-torque function using our custom calibration process. To do this, we fix an actuator to a 3D-printed fixture connected to a standard electronic scale, measure the torques by giving a wide range of current values, and create a linear interpolation mapping between current and torque values. We extrapolate to estimate values beyond the available data range to compute the torque.

\subsection{Actuator placement and transmission}

\subsubsection{Low inertia actuator placement}

\begin{figure}[hbt]
    \centering
    \includegraphics[width=0.57\linewidth]{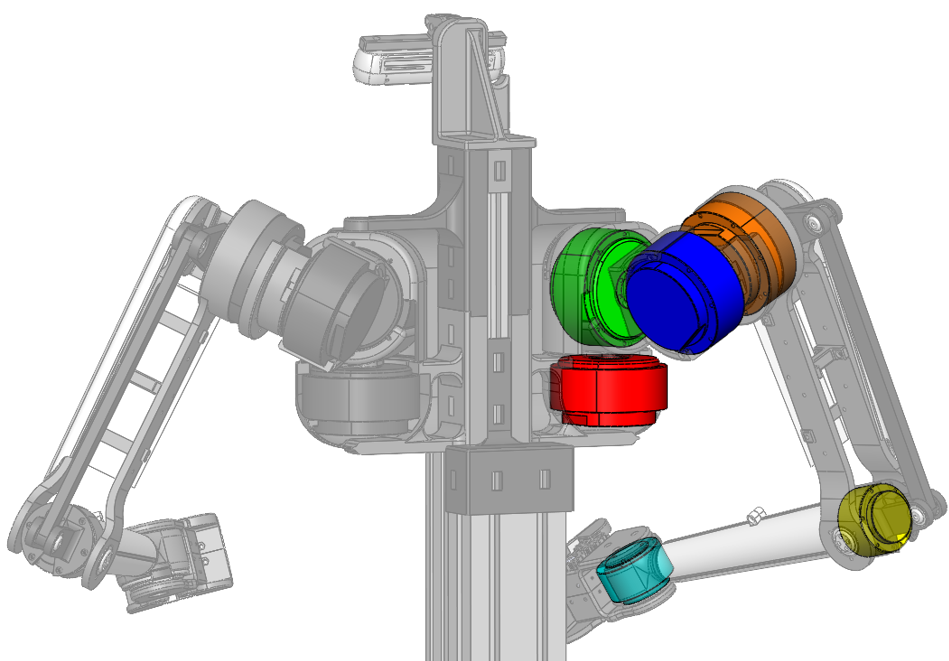}
    \caption{Positions of the six actuators. Four heavy and strong actuators are attached to the base. Two small actuators to rotate the wrist is attached to the elbow and wrist.}
    \label{fig:dof_position_back}
\end{figure}

\robot arm features 6 DoF: three on the shoulder, one on the elbow, and two on the wrist. We place the shoulder and elbow actuators near the base as illustrated in Figure~\ref{fig:dof_position_back}. By positioning the heavy actuators near the shoulder and transmitting the motion via the linkage system, we can effectively reduce the moment of inertia of the arm, achieving faster motion with the same amount of power from the actuators. We attach the two weaker but lightweight actuators directly on the elbow and wrist to keep the design simple.

\subsubsection{Parallelogram linkage for one-to-one torque transfer}
\label{sec:linkage}

One of the body-mounted actuators drives the elbow joints through a four-bar parallelogram linkage, as illustrated in Figure~\ref{fig:arm_assembly_and_gripper}a. This linkage transfers torque in 1:1 ratio, preserving the low gear ratio of the actuator and simplifying the model used in simulation and control.


\subsection{Material selection}
We primarily build the arm using 3D printing with PLA. This choice of material provides several advantages. First, compared to metals, which are the common choice for existing robot arms, PLA greatly reduces the total mass of the arm. Second, 3D printing is cost-effective and enables rapid part replacement for design iteration and maintenance. One disadvantage of PLA is that it is non-rigid compared to metals. This may harm the durability of the hardware and reduce control accuracy when exposed to high loads. Therefore, for the linkage for the elbow, where high loads are applied, we use aluminum, as shown in Figure~\ref{fig:arm_assembly_and_gripper}a. This way, we minimize the complicated and costly machining process while maintaining the rigidity sufficient for manipulation. To validate that our robot attains enough rigidity, we perform finite element method (FEM) analysis. Specifically, we assess the arm's deformation under a 10 N load applied along the x-axis (leftward) and the z-axis (downward).

\begin{figure}[hbt]
    \centering
    \includegraphics[width=0.75\linewidth]{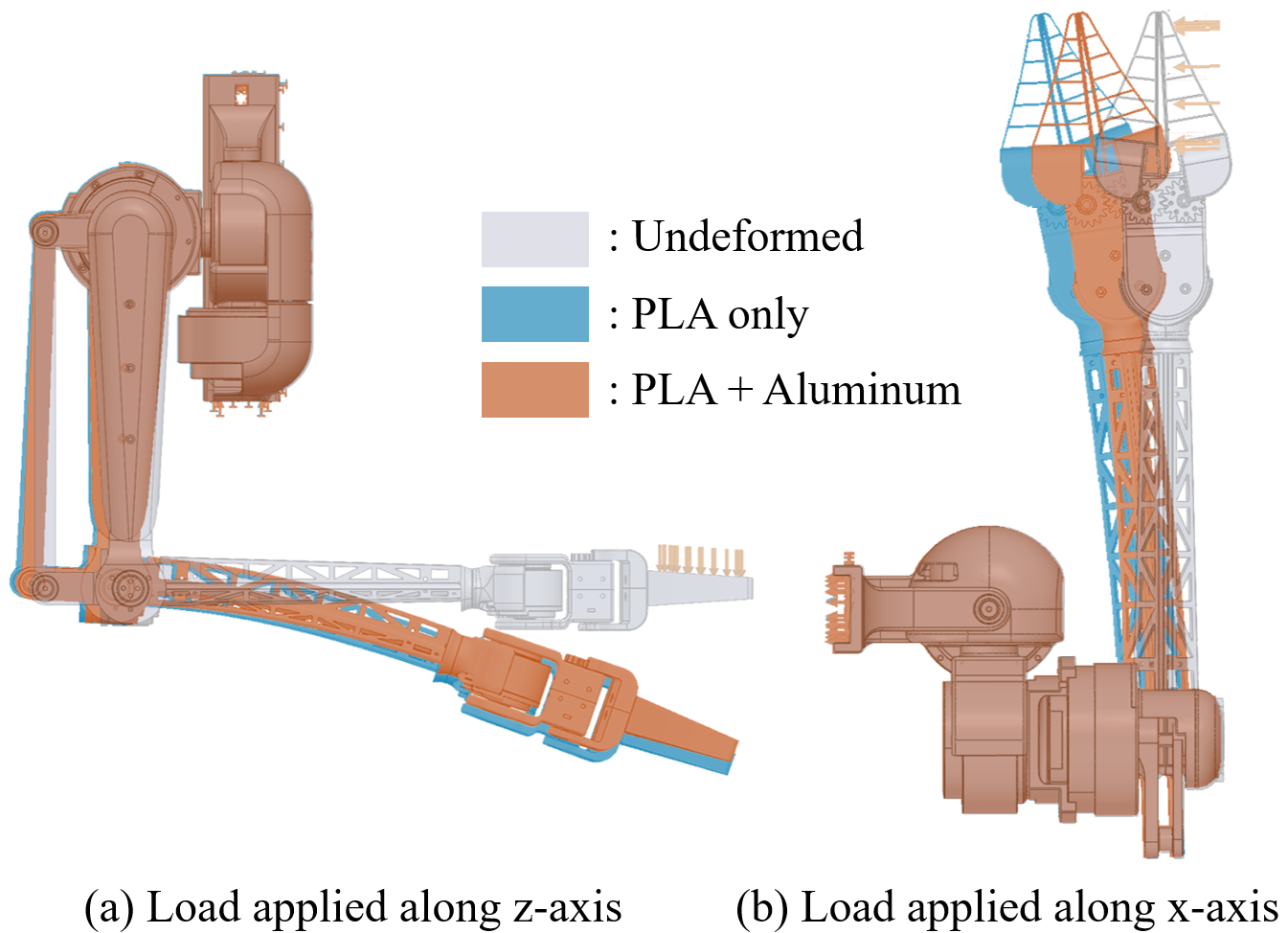}
    \caption{FEM analysis of deformation under a 10 N load applied along the (a) z-axis and (b) x-axis for configurations with all PLA and partially aluminum components. Deformation reduces from 1.69 mm to 1.58 mm (z-axis) and 2.85 mm to 2.24 mm (x-axis) with aluminum reinforcement. Deformation is exaggerated for clarity and visualization.}
    \label{fig:analysis}
\end{figure}

Figure~\ref{fig:analysis} illustrates the deformation results under z-axis and x-axis loading, respectively. Although this material change involves a trade-off with an increase in moving mass from 0.962 kg to 1.09 kg, it reduces deformation under loading conditions. For z-axis loading, the deformation decreases from 1.69 mm with PLA components to 1.58 mm with the addition of aluminum. Similarly, for x-axis loading, the deformation is reduced from 2.85 mm to 2.24 mm.

However, even at low loads, inaccuracies in the four-bar linkage components, especially in the rotational alignment of bearings and shafts, can significantly increase the end-effector position error. While PLA offers lighter weight and easier replacement, 3D-printed parts lack the dimensional accuracy needed for consistent performance, compared to CNC machining. Therefore, despite the challenges posed by aluminum part replacement, we selectively use them in critical regions to preserve precision.


\subsection{Custom gripper}
\begin{figure}
    \begin{subfigure}[t]{\linewidth}
        \centering
        \includegraphics[width=0.85\linewidth]{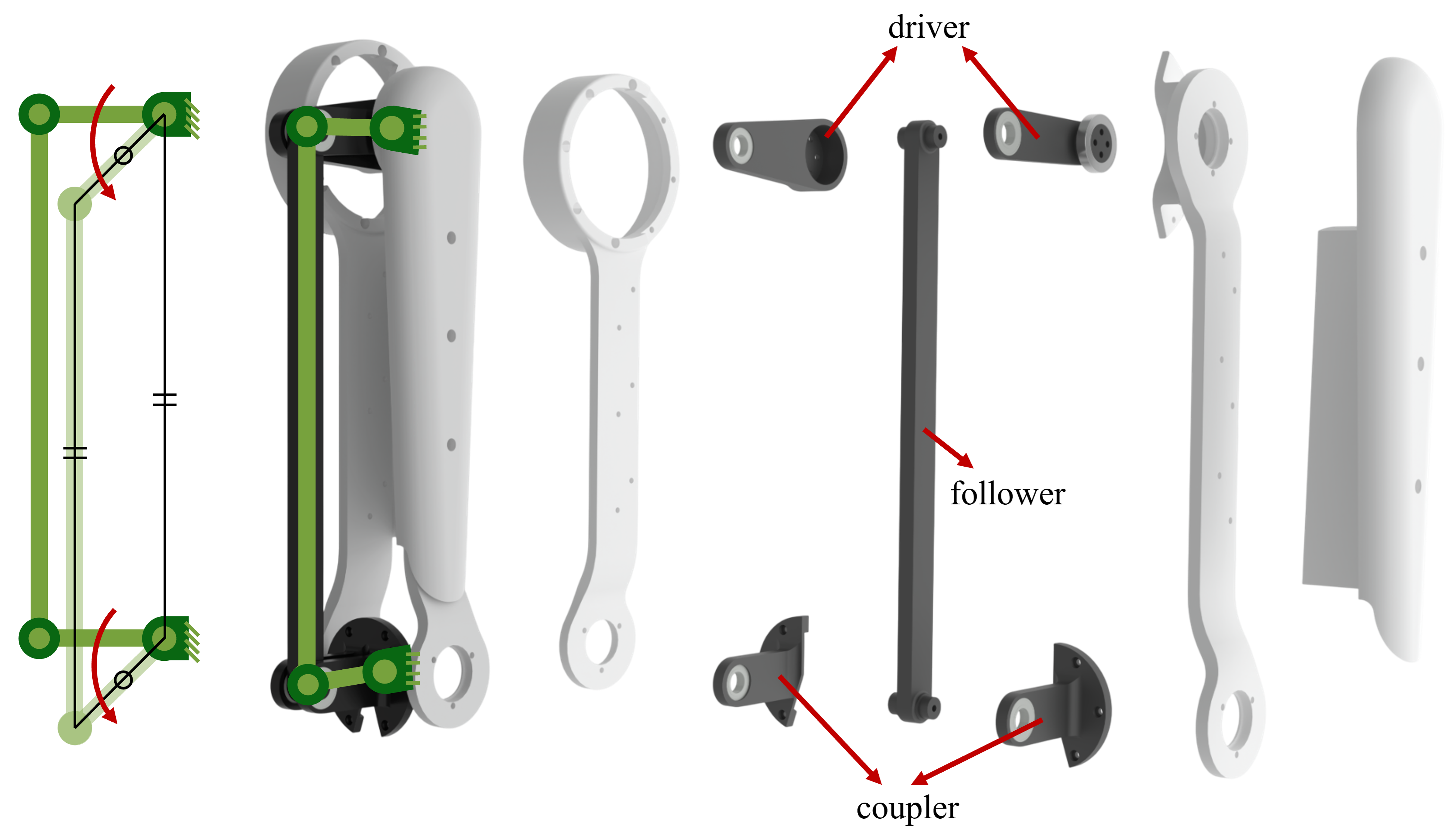}
        \caption{Arm assembly}
    \end{subfigure}
    \begin{subfigure}[t]{\linewidth}
        \centering
        \includegraphics[width=0.8\linewidth]{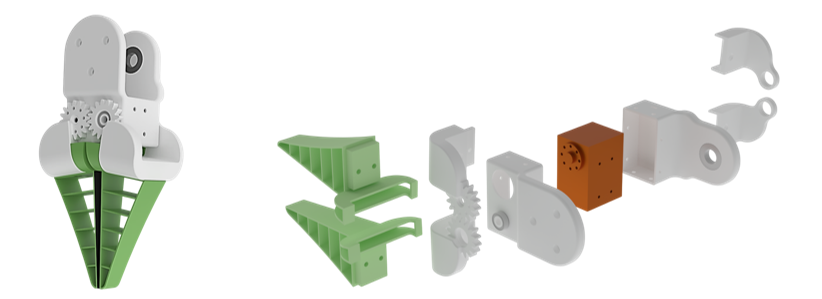}
        \caption{Gripper}
    \end{subfigure}
    \caption{(a) Assembly of the upper arm. Black parts are aluminum-machined, and white parts are 3D printed PLA. (b) Assembly of the custom-designed gripper. The brown component is the Dynamixel XM430 motor, the white parts are PLA parts, and the green parts are flexible TPU fingers.}
    \label{fig:arm_assembly_and_gripper}
\end{figure}

We also design a compact jaw gripper driven by a Dynamixel XM430 motor. The gripper utilizes a simple direct-drive 1 DoF gear mechanism, enabling jaw manipulation while maintaining ease of assembly. The gear mechanism is made of PLA, while the gripper pads are printed from flexible TPU to accommodate objects of various shapes. The entire assembly weighs approximately 290 g. Figure~\ref{fig:arm_assembly_and_gripper}b illustrates the design of the gripper.

\section{Mechanical Analysis}

\subsection{End-effector speed}

\begin{figure}
    \begin{subfigure}[b]{0.47\linewidth}
        \centering
        \includegraphics[width=0.92\linewidth]{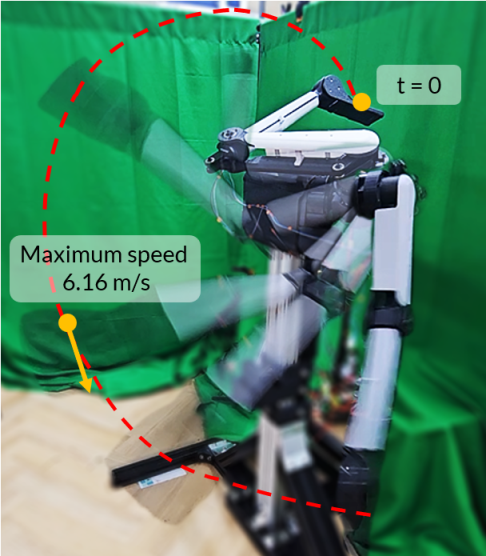}
        \caption{Speed test}
    \end{subfigure}
    \begin{subfigure}[b]{0.471\linewidth}
        \centering
        \includegraphics[width=\linewidth]{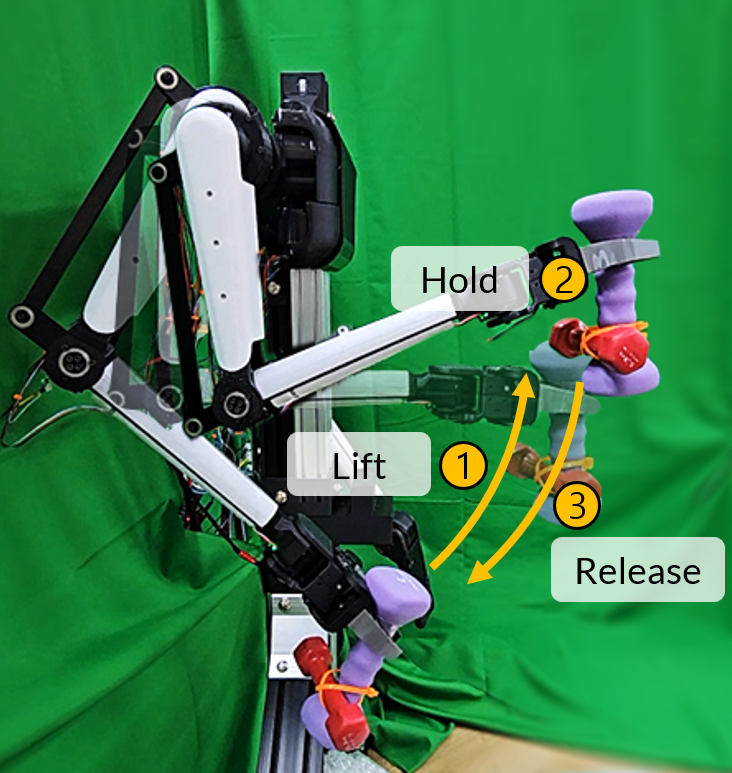}
        \caption{Lifting test}
    \end{subfigure}
    \begin{subfigure}[t]{\linewidth}
        \centering
        \includegraphics[width=0.99\linewidth]{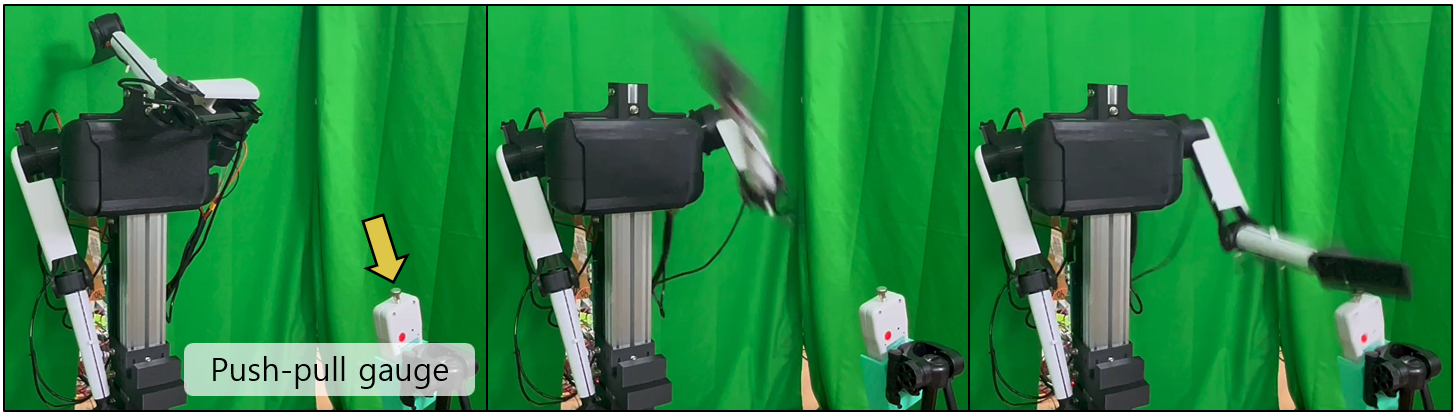}
        \caption{Impact test}
    \end{subfigure}
    \caption{(a) Trajectory of the end-effector speed test. Starting from the initial joint position, \robot accelerates toward the terminal position. (b) A sequence of a realistic dumbbell lifting task. The robot grasps, lifts, holds for three seconds, and places down the dumbbell. (c) We use push-pull gauge to measure the maximum impact force near the point of maximum speed.}
    \label{fig:speed_payload_impact}
\end{figure}

To test whether \robot can perform dynamic manipulation, we measure its end-effector maximum speed. To do this, we initialize the robot with its right end-effector positioned at the opposite shoulder with the elbow completely flexed, and then fully extend the elbow downward, as shown in Figure~\ref{fig:speed_payload_impact}a. The joint trajectory is generated by interpolating between the initial and final configurations, and joint position control is used to follow the trajectory. \robot achieves an average maximum end-effector speed of 6.16 m/s with a standard deviation of 0.472 m/s across 40 repetitions, without any damage to the arm or power supply.

\subsection{Impact force at the maximum velocity}
To evaluate the safety of \robot, we measure its impact force near maximum velocity using a push-pull gauge, as shown in Figure~\ref{fig:speed_payload_impact}c. Over eight repetitions, \robot records an average impact force of 50.5 N, with a maximum of 52.2 N and a standard deviation of 5.55 N. For comparison, a typical human palm strike at a similar speed generates approximately 575 N of impact force, based on an average effective hand mass of 1.39 kg~\cite{adamec2021biomechanical} and an impact duration of 13 ms~\cite{walilko2005biomechanics}, which is nearly 10 times that of \robot. While humans have soft skin and \robot does not, this demonstrates \robot can operate safely even at its maximum speed.



\subsection{Payload}

We evaluate \robot's payload using dumbbells, as shown in Figure~\ref{fig:speed_payload_impact}b. The test involves performing bicep curls with the dumbbell.
By gradually increasing the weights, we find that \robot can hold up to 2.5 kg, compared to 3 kg for existing collaborative robots. The arm and gripper reliably grasp, lift, hold, and release the weight across ten consecutive trials without any damage or failure.

\begin{figure}[hbt]
\begin{subfigure}{\linewidth}
    \centering
    \includegraphics[width=0.7\linewidth]{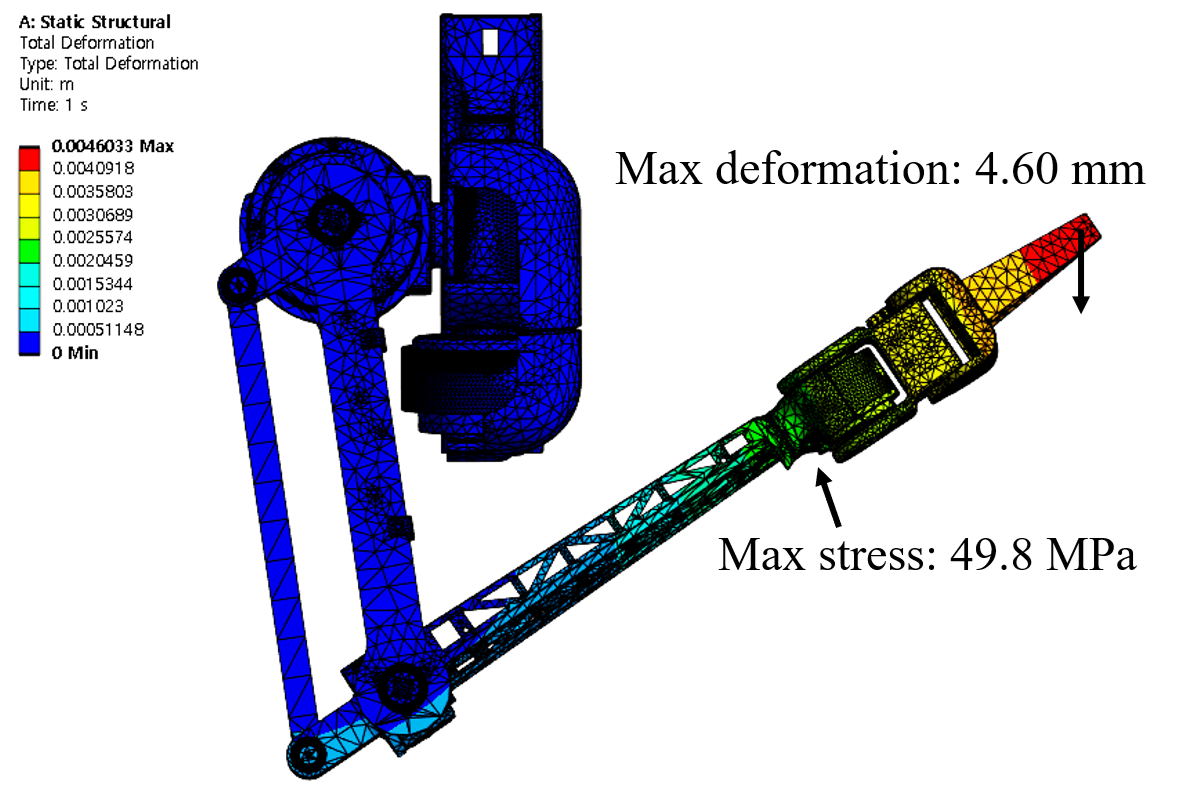}
    \caption{FEM result}
\end{subfigure}
\vfill
\begin{subfigure}{\linewidth}
    \centering
    \includegraphics[width=0.95\linewidth]{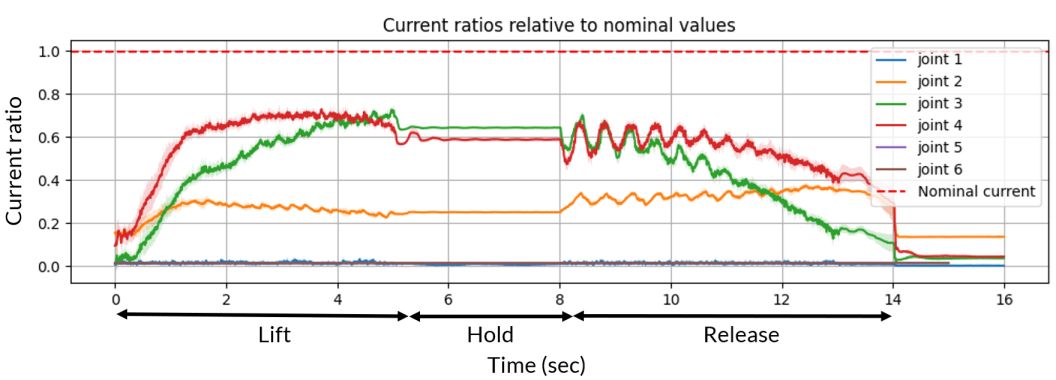}
    \caption{Current graph during payload experiment}
\end{subfigure}
\caption{(a) FEM analysis under a 2.5 kg payload in position 2 in Figure~\ref{fig:speed_payload_impact}b. Maximum deformation occurs at the gripper tip, measuring 4.60 mm, while the maximum stress is observed at the wrist link at 49.8 MPa. Note that the material of the TPU-based gripper is replaced with PLA for FEM analysis, as it is difficult to accurately simulate TPU's flexibility. (b) The average ratio of actuator current to its nominal value for each actuator and their standard deviations, measured during ten repetitions under a 2.5 kg payload.} 
\label{fig:payload}
\end{figure}

We analyze both mechanical stress and current to evaluate the reliability under the maximum payload. To measure mechanical stress, the robot configuration is set to position 2 in Figure~\ref{fig:speed_payload_impact}b. Each link’s stress limit is determined by its tensile strength, which is defined as the material’s resistance to breaking under tension. Using FEM analysis in Ansys, as shown in Figure~\ref{fig:payload}a, we find that the maximum stress experienced by the arm while lifting a 2.5 kg payload is 49.8 MPa, which is within the tensile strength of the PLA filament, 61 MPa. This indicates that \robot would deform, but not break and can safely operate under its maximum payload without significant risk of structural failure.

Each actuator has a nominal current limit, the maximum current it can handle without overheating. To evaluate whether the robot operates safely within this limit, we measure the input current throughout the entire payload test including lifting, holding, and releasing. Figure~\ref{fig:payload}b shows the current ratios relative to nominal values for each actuator. We can see that all currents are under their nominal values.

\begin{figure}[hbt]
    \centering
    \includegraphics[width=0.5\linewidth]{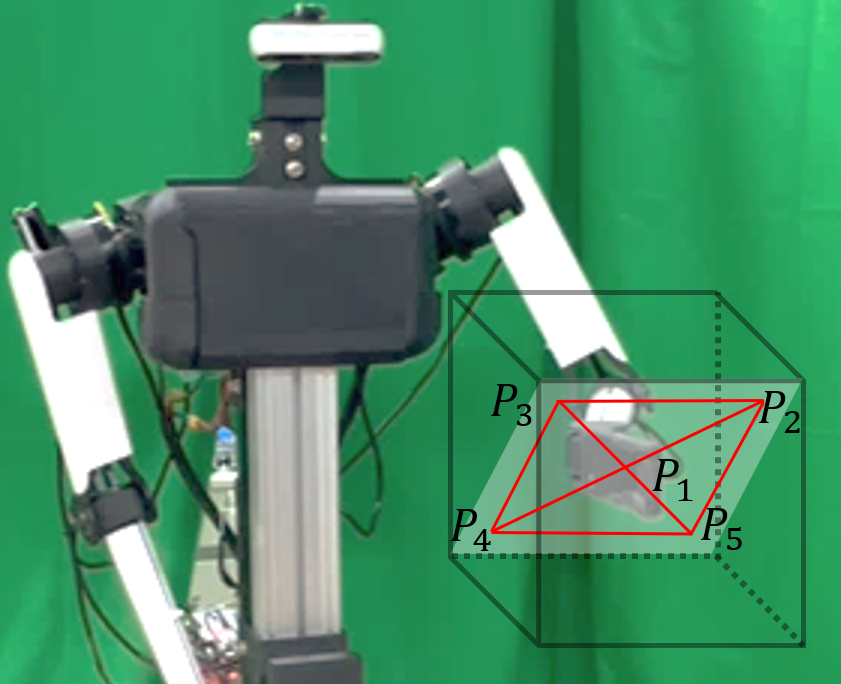}
    \caption{Repeatability test setup with five end-effector points $P_1$ to $P_5$. The test plane is placed diagonally within the 250 mm cube. $P_1$ is positioned at the center of the plane, while $P_2$, $P_3$, $P_4$, and $P_5$ are near the corners, each located one-tenth of the cube's diagonal length from the center.}
    \label{fig:repeatability}
\end{figure}

\subsection{Repeatability}
To evaluate \robot's repeatability, we follow ISO 9283 standard~\cite{ISO9283:1998} to setup an experiment where the robot has to follow a designated set of points in a 250mm cube within \robot's workspace. Figure~\ref{fig:repeatability} shows a test plane placed diagonally within the cube. Five end-effector points $P_1$ to $P_5$ are marked on the plane, with $P_1$ at the center and $P_2$, $P_3$, $P_4$, and $P_5$ near the corners, each located one-tenth of the cube's diagonal length from the center. \robot sequentially moves through these five points and repeats the trajectory 30 times with its end-effector facing forward. We use 12 OptiTrack cameras to monitor it simultaneously to record the end-effector's position. For each point, \( N=30 \) is the total number of measurements, and  $P_i\in \mathbb{R}^3, i=1, 2, \cdots, N $ is the measured position at the $i^{th}$ instance. The ISO 9283 standard states that the mean position of the end-effector $\bar{P} = \frac{1}{N} \sum_{i=1}^N P_i $ is calculated as the average of the measured positions $P_i$. The repeatability $R$ is computed based on the average (\( \mu \)) and the standard deviation (\( \sigma \)) of the distances between \( \bar{P} \) and $P_i$.

\begin{equation*}
    \mu = \frac{1}{N} \sum_{i=1}^N \lVert P_i - \bar{P} \rVert, \quad 
    \sigma = \sqrt{\frac{1}{N-1} \sum_{i=1}^N (\lVert P_i - \bar{P} \rVert)^2}
\end{equation*}

\begin{equation*}
    R = \mu + 3\sigma
\end{equation*} 

As shown in Table~\ref{tab:repeatability_results}, \robot achieves an average repeatability of 2.63 mm which means it is less consistent compared to the cobots with high gear ratio actuators, whose repeatability is around 0.1mm. This shows the trade-off between compliance and consistency.

\begin{table}[hbt]
\centering
\caption{ISO 9283 repeatability experiment results}
\label{tab:repeatability_results}
\resizebox{0.46\textwidth}{!}{%
\begin{tabular}{|c|c|c|c|}
\hline
Point & Average distance (mm) & Std dev. (mm) & Repeatability (mm)\\ \hline
$P_1$    & 1.048     & 0.546    & 2.687\\ 
$P_2$    & 1.042     & 0.409    & 2.269\\ 
$P_3$    & 0.939     & 0.689    & 3.006\\ 
$P_4$    & 0.751     & 0.520    & 2.311\\ 
$P_5$    & 1.104     & 0.584    & 2.857\\ \hline
Average    & 0.977     & 0.550    & 2.626\\ \hline

\multicolumn{3}{|c|}{Franka Panda~\cite{haddadin2022franka}} & 0.1 \\ 
\multicolumn{3}{|c|}{KUKA iiwa 7 R800} & 0.1 \\ 
\multicolumn{3}{|c|}{Quigley et al.~\cite{quigley2011low}} & *3 \\ 
\multicolumn{3}{|c|}{LIMS~\cite{kim2017anthropomorphic}} & 0.43 \\ 
\multicolumn{3}{|c|}{Nishii et al.~\cite{nishii2023ultra}} & *2.2 \\ 
\multicolumn{3}{|c|}{BLUE~\cite{gealy2019quasi}} & *3.7 \\ \hline
\end{tabular}%
}
\parbox{0.46\textwidth}{
\vspace{2mm}
\textit{``*"} indicates that repeatability is obtained manually, not by ISO 9283.

}
\end{table}



\section{Experiments}

\subsection{Dynamic manipulation}
We conduct three experiments, snatching, hammering, and batting, to show that \robot can perform dynamic manipulation. In these experiments, the target object pose is known in advance, and a pre-defined trajectory is executed with a joint position control to focus on demonstrating raw hardware capability.


\begin{figure}[hbt]
    \begin{subfigure}{\linewidth}
        \centering
        \includegraphics[width=0.7\linewidth]{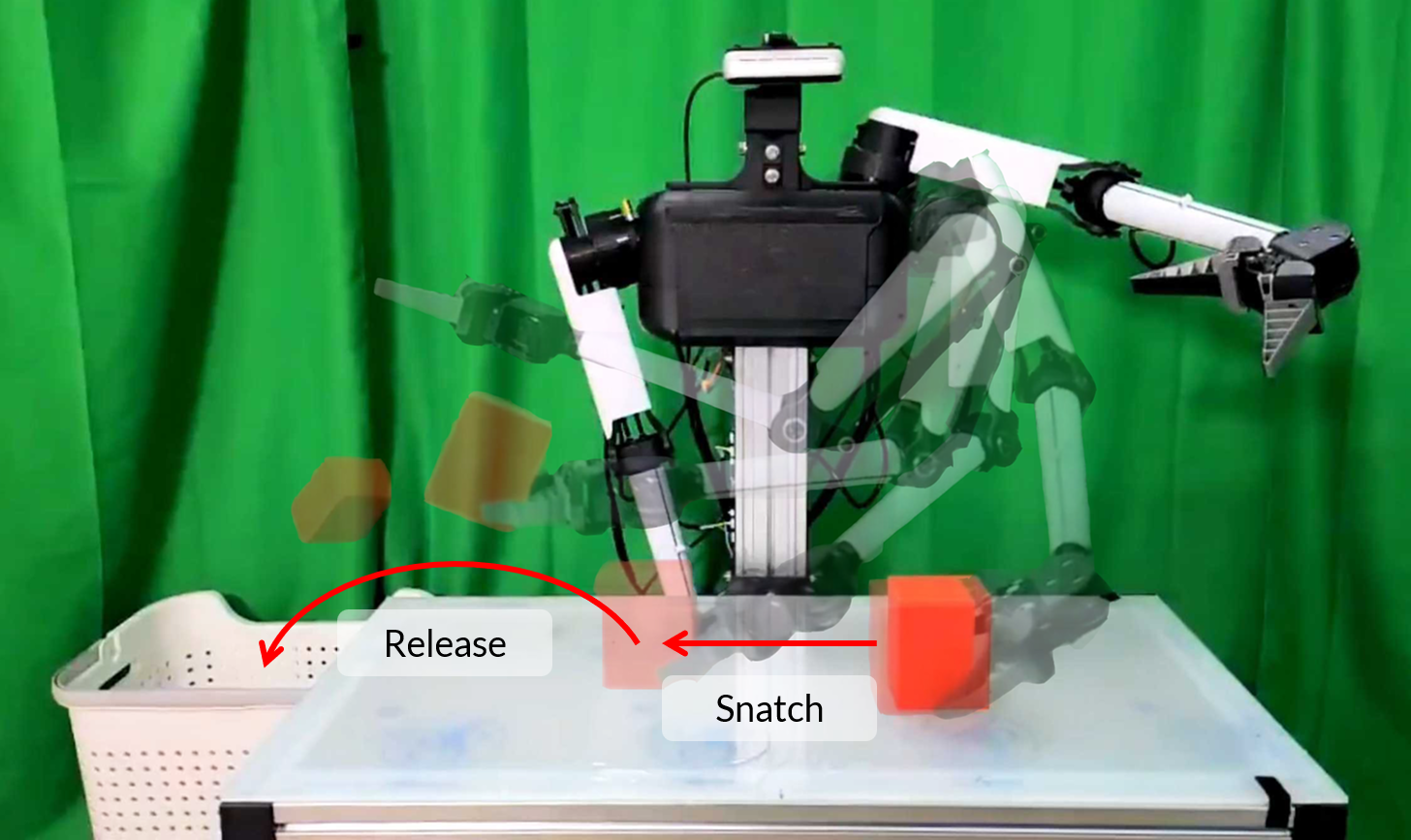}
        \caption{Experiment setup}
    \end{subfigure}
    \vfill
    \begin{subfigure}{\linewidth}
        \centering
        \includegraphics[width=0.6\linewidth]{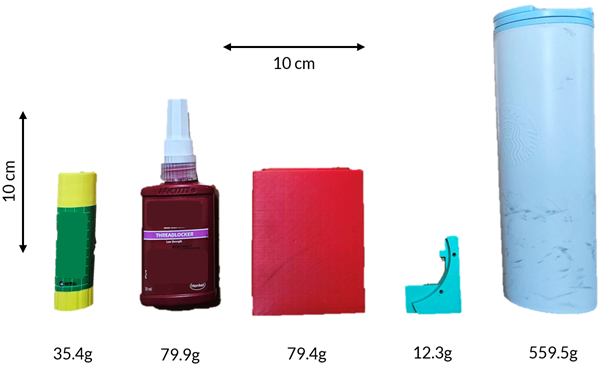}
        \caption{Objects for snatching}
    \end{subfigure}
    \caption{(a) Snatching experiment setup. The object is placed in a fixed pose, and \robot rapidly snatches and releases it into the white basket. (b) Various real-world objects used in snatching which have different sizes and masses.}
    \label{fig:snatching}
\end{figure}

\subsubsection{\textbf{Snatching}}
Unlike static pick-and-place, snatching involves swift single-arm manipulations, where the robot rapidly approaches, grasps, and releases the target object. The experiment setup is shown in Figure~\ref{fig:snatching}a, where the goal is to snatch the object and releases it into a basket beside the table. We test \robot with five objects of varying shapes and masses, with ten trials for each object. Detailed information on the sizes and masses of these objects is described in Figure~\ref{fig:snatching}b. \robot achieves an overall success rate of 80\%, successfully completing all trials and objects except for the heavy tumbler, as shown in Table~\ref{tab:snatching_result}. This highlights the strength and limitation of \robot, and the difficulty of snatching. While it can snatch light-weight objects, it often drops the heavy tumbler (560 g) after snatching. Such heavy object demands a much more delicate yet firm grasp to counteract momentum and gravity effectively. Please see our supplement videos on snatching for better visualization.


\begin{table}[hbt]
\centering
\caption{Snatching experiment results}
\label{tab:snatching_result}
\resizebox{0.99\columnwidth}{!}{%
\begin{tabular}{|c|c|c|c|c|c|c|}
\hline
Object          & Glue & Threadlocker & Plastic box & Bracket & Tumbler & Total  \\ \hline
Success/Trial & 10/10   & 10/10  & 10/10  & 10/10  &  0/10   & 40/50 \\ \hline
\end{tabular}%
}
\end{table}

\subsubsection{\textbf{Hammering}}
\begin{figure}[hbt]
    \centering
    \begin{subfigure}[b]{0.425\linewidth}
        \centering
        \includegraphics[width=\linewidth]{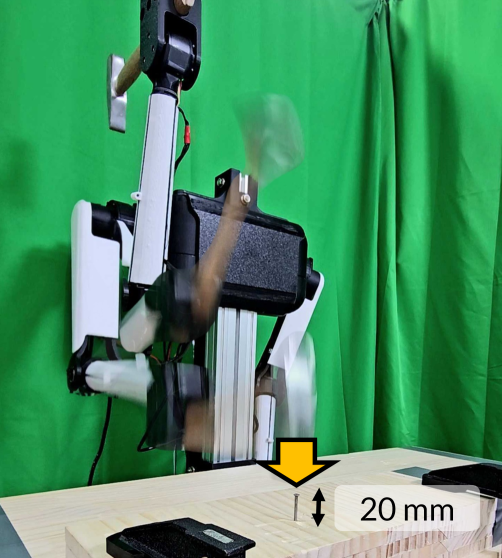}
        \caption{batting experiment setup}
    \end{subfigure}
    \begin{subfigure}[b]{0.5\linewidth}
        \centering
        \includegraphics[width=\linewidth]{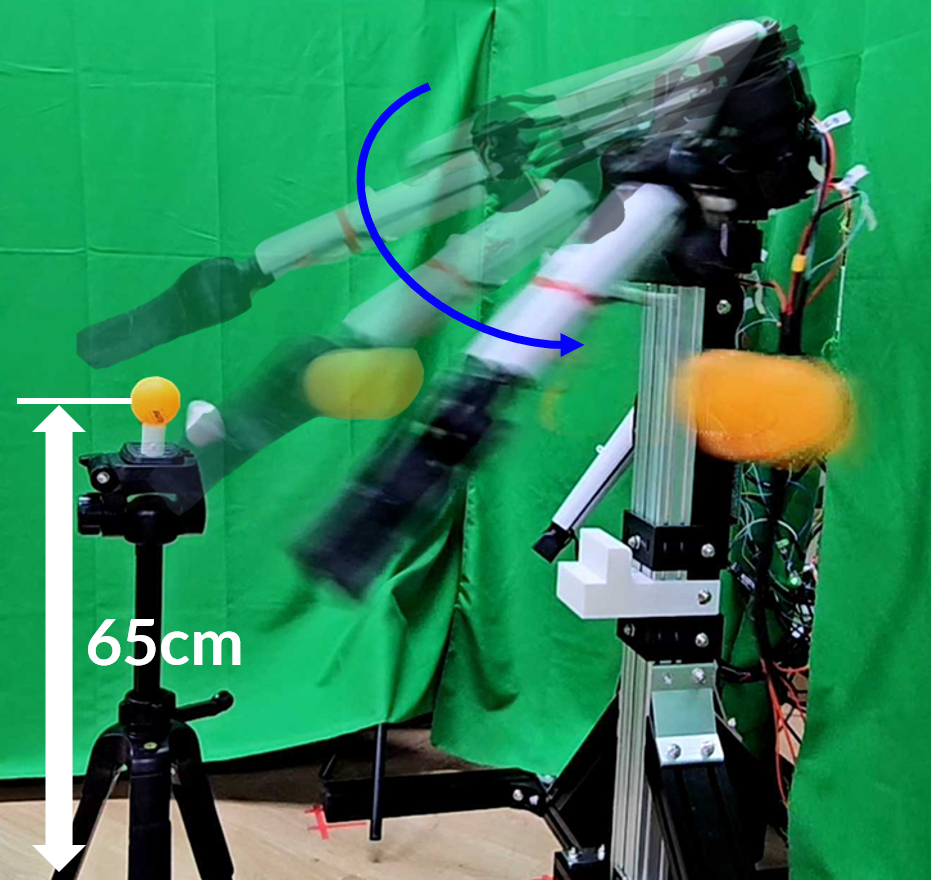}
        \caption{card domain setup}
    \end{subfigure}
    \caption{(a) Hammering experiment setup. A nail is fixed on a wooden board with a height of 20 mm. The robot repeatedly strikes the nail until fully driven into the board. (b) Batting experiment setup. The ping pong ball is placed 65 cm above the ground beside ARMADA.}
    \label{fig:hammering_batting}
\end{figure}

This task involves repeatedly striking a nail into a wooden board to evaluate the robot's capability of creating significant impact force. The experiment setup is shown in Figure~\ref{fig:hammering_batting}a. We record the number of strikes required to drive the nail 20 mm into the wooden board. If the nail bends during hammering, a human intervenes and straightens it with a wrench. For comparison, 14 humans (2 females and 12 males) perform the same single-hand hammering task under identical conditions, using the same wooden board, nail, and hammer. For human participants, it takes an average $13.15$ strikes with a standard deviation of $6.41$ to accomplish the task. On the other hand, \robot achieves a comparable level of efficiency to humans, driving the nail with an average of $10$ strikes with a standard deviation of $1.26$. Please see our supplement video on hammering to better grasp the speed and impact that \robot can generate.




\subsubsection{\textbf{Batting}}

Batting requires sufficient impact on a projectile in a short period of time. As illustrated in Figure~\ref{fig:hammering_batting}b, ARMADA swings its arm with a pre-computed trajectory to hit the ping pong ball with the dorsum of its end-effector~(EE). We record the distance between the base of the ball stand and the point where the ball touches the ground first. We repeat the batting 20 times, and the average driving distance of the ball is $\textbf{3.484~m}$ with a standard deviation of 0.385~m. The average speed of the EE at the impact is 6.135~m/s. Since the length of ping pong table is 2.74~m, ARMADA can produce enough impact to play ping pong.

\subsection{RL-based non-prehensile manipulation}
To validate our claim that \robot can be used for RL, we train \robot on a contact-rich non-prehensile manipulation task entirely in simulation and zero-shot transfer the policy to the real world. We demonstrated two tasks, each named as the bump and the card task.

\subsubsection{\textbf{Bump}}
\begin{figure}[hbt]
    \centering
    \includegraphics[width=0.99\linewidth]{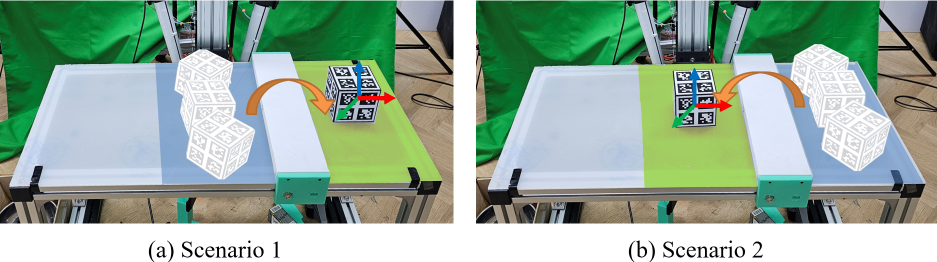}
    \caption{(a) Scenario 1: The robot moves the object from the right to the left of the bump. (b) Scenario 2: The robot moves the object from the left to the right. The object's initial pose is randomly set within the blue area. The robot must manipulate the object to match both the position and orientation of the goal pose sampled within the green area.}
    \label{fig:bump_real}
\end{figure}

In this bump task, the robot uses a single arm to manipulate a cube whose length is 90mm over a 25 mm bump obstacle on the table, as shown in Figure~\ref{fig:bump_real}. Starting from a random initial pose, the robot must push, topple, strike, and reorient the object to overcome the bump and put it at the specified goal position and orientation without dropping it. We train the policy using NVIDIA Isaac Gym~\cite{makoviychuk2021isaac} with extensive domain randomization and zero-shot transfer to the real robot. Details regarding the training, including hyperparameters and MDP definitions are in Appendix~\ref{app:rl_detail}.

We perform 20 trials in total, 10 on each scenario. \robot achieves a 9/10 success rate in scenario 1, with a single failure where the cube falls off the table during the manipulation. In scenario 2, it achieves 7/10, where failures occur because the robot fails to make solid contact with the object. Most of the errors happen because of the sim-to-real gap in the environment and object, such as the object friction coefficient. Overall, the robot demonstrates 80\% success rate across both scenarios. 

\subsubsection{\textbf{Card}}

\begin{figure}[hbt]
    \centering
    \includegraphics[width=0.4\linewidth]{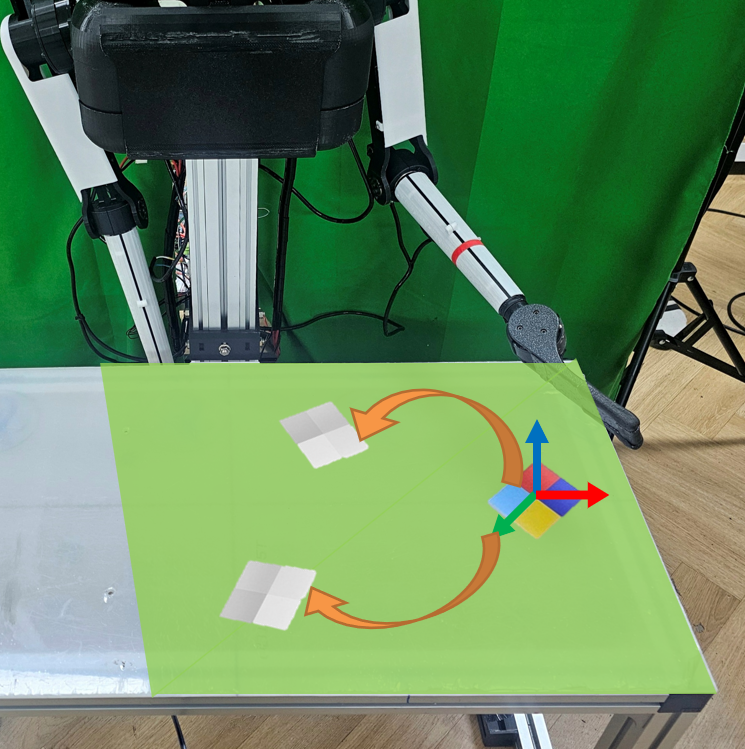}
    \caption{In the card domain, the object's initial and goal poses are randomly set within the green area.}
    \label{fig:card_real}
\end{figure}

Figure~\ref{fig:card_real} illustrates the domain where the robot uses a single arm to manipulate a card too flat to be grasped directly. The card is placed at a random pose on the table. The robot pushes, drags, and reorients the card to the specified goal pose. We train everything in simulation and zero-shot transfer to the real world. Out of 20 trials, ARMADA achieves a success rate of $\textbf{17/20}$. Failures are mostly due to sim-to-real gap such as the friction coefficient of the table and the card.

\subsection{Human motion shadowing: Bimanual throwing}
In this task, we re-target human joint poses to the robot to show that \robot can serve as the platform for learning dynamic motions from a human demonstration. We use WHAM~\cite{wham} to predict SMPL~\cite{smpl} parameters, which estimate human joint keypoints, joint angles, and body shape to describe the pose of the human body. However, simply copying the joint angles to the corresponding actuators results in inaccurate motion due to the discrepancy of morphology between the SMPL model and \robot. We instead track the positions of the elbow and wrist joints of the human operator and solve differential inverse kinematics using PINK~\cite{pink} to find the joint configuration that minimizes the distance between the robot and the human joints at the elbow and wrist. This enables \robot to mimic human motion while preserving the relative position between the two hands, which is crucial in a bimanual task. 

\begin{table}[hbt]
\centering
\caption{Box driving distance}
\label{tab:box_distance}
\resizebox{0.9\columnwidth}{!}{%
\begin{tabular}{|c|c|c|c|c|c|c|c|}
\hline
Trial        & 1    & 2    & 3    & 4    & 5    & Avg. & Std dev.   \\ \hline
Distance (m) & 1.80 & 2.07 & 1.81 & 2.20 & 2.25 & 2.03 & 0.19 \\ \hline
\end{tabular}%
}
\end{table}

To show that \robot can shadow dynamic motion, we perform bimanual throwing task where the human shows a quick motion for throwing a box, and the robot's goal is to mimic that behavior to throw a box (318 g) placed 515 mm above the ground and 300 mm in front of its base as quickly as possible. The overall process of throwing is shown in Figure~\ref{fig:overall_illustration}d. Since the driving distance of the box is directly proportional to the end-effector speed at the moment of release, we measure the distance between the base and the closest part of the box. We repeat the experiment 5 times. \robot throws the box 2.03 m on average, as shown in Table~\ref{tab:box_distance}, and during throwing, \robot achieves an average maximum speed of 3.56 m/s with the left hand and 3.92 m/s with the right hand. To view the more detailed motion of \robot, please see our supplement video on bimanual throwing.

\section{Limitation}
The use of 3D-printed PLA for structural components improves ease of assembly and reduces weight and cost, yet it causes deformation under heavy load, which can diminish end-effector precision. Using metal, such as aluminum, would remedy this problem. Additionally, \robot relies on integrated joint relative encoders, requiring manual initialization in a fixed joint configuration each time the system is powered on. Using absolute joint encoders could significantly improve accuracy and ease of use, although it would increase the overall cost. 



The 6 DoF configuration of the arm provides sufficient mobility for single-arm manipulation tasks, yet it shows a limitation in certain bimanual manipulation problems. Specifically, when \robot holds onto a rigid object with both hands, each arm loses 1 DoF because the hands are fixed to the object during grasping. This leads to an underactuated kinematic chain which has a limited mobility in 3D space. We can achieve more mobility by letting the object slip inside the grippers, yet this renders the grasp less robust and simulation difficult. Therefore, we anticipate that designing a lightweight 3 DoF wrist in place of the current 2 DoF wrist allows a more diverse repertoire of manipulation in bimanual tasks.

Finally, the limited torque density of commercially available proprioceptive actuators restricts the performance. Currently, all of our actuators feature a 1:10 gear ratio, so \robot can handle up to 2.5 kg of payload. To handle a heavier object and manipulate it with higher torque, we expect the actuator to have 1:20$\sim$30 gear ratio, but it is difficult to find an off-the-shelf product that meets our requirements. Customizing the actuator to increase the torque density while minimizing the weight will enable \robot to move faster and handle more diverse objects.


\section*{Acknowledgments}

This work was supported by Institute of Information \& communications Technology Planning \& Evaluation (IITP) grant and National Research Foundation of Korea (NRF) funded by the Korea government(MSIT) (No.2019-0-00075, Artificial Intelligence Graduate School Program(KAIST)), (No.2022-0-00311, Development of Goal-Oriented Reinforcement Learning Techniques for Contact-Rich Robotic Manipulation of Everyday Objects), (No. 2022-0-00612, Geometric and Physical Commonsense Reasoning based Behavior Intelligence for Embodied AI), (No. RS-2024-00359085, Foundation model for learning-based humanoid robot that can understand and achieve language commands in unstructured human environments), (No. RS-2024-00509279, Global AI Frontier Lab). We would also like to thank Prof. Haewon Park for his great guidance.


\bibliographystyle{plainnat}
\bibliography{references}

\section*{Appendix}




\subsection{Non-prehensile Manipulation Policy Details}
\label{app:rl_detail}

\subsubsection{Simulation Setup}

The RL training setup uses 24,576 simulated environments in NVIDIA IsaacGym, running on an NVIDIA GeForce RTX 4090 GPU and an AMD Ryzen 7 7800X3D CPU, with training completed in 2 days. Initial configurations are randomly set to collision-free configuration, without requiring an additional $\pi_{pre}$ to enforce contact between the robot's end-effector and the object. The policy operates at 20 Hz, while the controller runs at 200 Hz. Domain randomization (DR) is applied extensively to enhance robustness, including variations in table height, joint positions, velocities, end-effector position, object properties (friction and mass), and torque noise. The ranges of these parameters are shown in Table~\ref{tab:DR}. The environment design mirrors the bump and cube dimensions from Kim et al.~\cite{kim2023pre}, but with an enlarged table size of $84 \times 40$ cm to accommodate the bimanual workspace of our robot.

\begin{table}[ht]
\centering
\caption{Domain randomization parameters and their ranges. $\mathcal{U}[min, max]$ denotes uniform distribution, and $\mathcal{N}[\mu,\sigma]$ denotes Normal distribution.}
\label{tab:DR}
\resizebox{0.6\linewidth}{!}{%
\begin{tabular}{|c|c|}
\hline
\textbf{Parameter}           & \textbf{Range}                      \\ \hline
Table height                 & $+\mathcal{U}[-0.01, 0.01]$         \\ 
Object friction              & $\times\mathcal{U}[0.7, 1.3]$             \\ 
Object mass                  & $\times\mathcal{U}[0.7, 1.3]$             \\ 
Joint positions              & $+\mathcal{N}[0.0, 0.05]$           \\ 
Joint velocities             & $+\mathcal{N}[0.0, 0.05]$           \\ 
End-effector position        & $+\mathcal{N}[0.0, 0.05]$           \\ 
Torque noise                 & $+\mathcal{N}[0.0, 0.1]$            \\ \hline
\end{tabular}%
}
\end{table}

\subsubsection{Reinforcement Learning Setup}

The state space $\mathcal{S}$ includes joint positions and velocities, current and goal object keypoints, end-effector pose, and the previous action. Details of the state space components are provided in Table~\ref{tab:state_space}. The action space $\mathcal{A}$ comprises joint position residual, proportional gain, and derivative gain for each joint, as summarized in Table~\ref{tab:action_space}. The policy architecture follows $\pi_{post}$ from Kim et al.~\cite{kim2023pre}.

\begin{table}[ht]
\centering
\caption{Components of the state space $\mathcal{S}$.}
\label{tab:state_space}
\resizebox{0.8\linewidth}{!}{%
\begin{tabular}{|c|c|}
\hline
\textbf{Component}           & \textbf{Description}               \\ \hline
$q[t] \in \mathbb{R}^{6}$    & joint positions at time step $t$        \\ 
$\dot{q}[t] \in \mathbb{R}^{9}$ & joint velocities at time step $t$     \\ 
$u_{o}[t] \in \mathbb{R}^{2 \times 8}$ & 2D object key points at time step $t$       \\ 
$u_{g} \in \mathbb{R}^{2 \times 8}$ & 2D goal object key points                      \\ 
$T_{E}[t] \in SE(3)$ & end-effector pose at time step $t$                   \\ 
$a_{post}[t-1] \in \mathbb{R}^{18}$ & action at time step $t-1$              \\ \hline
\end{tabular}%
}
\end{table}

\begin{table}[ht]
\centering
\caption{Components of the action space $\mathcal{A}$.}
\label{tab:action_space}
\resizebox{0.7\linewidth}{!}{%
\begin{tabular}{|c|c|}
\hline
\textbf{Component}           & \textbf{Description}               \\ \hline
$\Delta_q[t] \in \mathbb{R}^{6}$          & joint position residual                        \\ 
$k_p[t] \in \mathbb{R}^{6}$         & proportional gain for each joint             \\ 
$k_d[t] \in \mathbb{R}^{6}$         & derivative gain for each joint              \\ \hline
\end{tabular}%
}
\end{table}

\subsubsection{Real-World Experiment Setup}

The real-world experiment setup includes one Intel RealSense D455 and one D435 camera, calibrated using four AprilTags for accurate pose alignment. To ensure reliable object tracking, 24 AprilTags are distributed across the cube, with four tags on each face. This configuration maintains visibility even when some tags are occluded during manipulation.



\subsection{Bill of Materials}
\label{app:bom}
    
\begin{table}[ht]
\centering
\caption{Bills of materials for building a single arm including a gripper.}
\label{tab:bom}
\resizebox{\linewidth}{!}{%
\begin{tabular}{|c|c|c|c|}
\hline
Product name              & Quantity & Individual price (\$) & Set price (\$) \\ \hline
T-motor AK70-10           & 4                                                                  & 488.9                 & 1955.6         \\ \hline
RMD x4 v2 1:10            & 2                                                                  & 136.8                 & 273.5          \\ \hline
Robotiz XM430-W350-R      & 1                                                                  & 198.6                 & 198.6          \\ \hline
PLA filament 2kg          & 1                                                                  & 27.4                  & 27.4           \\ \hline
Mean Well SMPS LRS-600-24 & 2                                                                  & 43.5                  & 87.0           \\ \hline
MKS CANable Pro           & 3                                                                  & 8.5                   & 25.6           \\ \hline
SEMI-REX MD110-16 diode   & 2                                                                  & 13.7                  & 27.4           \\ \hline
Bearings                  &                                                                    & 13.7                  & 13.7           \\ \hline
Aluminum machining        &                                                                    & 425.3                 & 425.3          \\ \hline
Total price               &                                                                    &                       & 3034.0         \\ \hline
\end{tabular}%
}
\end{table}

\end{document}